\documentclass[10pt,twocolumn,letterpaper]{article}

 \usepackage[pagenumbers]{cvpr} 

\usepackage{algorithm}
\usepackage{algpseudocode}
\usepackage{amsmath}
\usepackage{float}

\definecolor{cvprblue}{rgb}{0.21,0.49,0.74}
\usepackage[pagebackref,breaklinks,colorlinks,allcolors=cvprblue]{hyperref}
\usepackage{fontawesome5}

\def\paperID{} 
\def\confName{CVPR}
\def\confYear{2026}

\title{APEX: A Decoupled Memory-based Explorer for Asynchronous Aerial Object Goal Navigation}

\author{
    Daoxuan Zhang$^{1}$\hspace{0.3cm}
    Ping Chen$^{1}$\hspace{0.3cm}
    Xiaobo Xia$^{2}$\hspace{0.3cm}
    Xiu Su$^{3}$\hspace{0.3cm}
    Ruichen Zhen$^{4}$\\
    Jianqiang Xiao$^{1}$\hspace{0.3cm}
    Shuo Yang$^{1\text{\faEnvelope}}$ \\
    \vspace{0.2cm}
    $^{1}$ Harbin Institute of Technology, Shenzhen\quad
    $^{2}$ National University of Singapore \\
    $^{3}$ Central South University\quad
    $^{4}$ Meituan Academy of Robotics Shenzhen, Meituan \\
    \vspace{0.2cm}
    \small \texttt{2023311529@stu.hit.edu.cn} \quad
    \small \texttt{shuoyang@hit.edu.cn} \\
}

\begin{document}
\maketitle
\begin{abstract}
Aerial Object Goal Navigation, a challenging frontier in Embodied AI, requires an Unmanned Aerial Vehicle (UAV) agent to autonomously explore, reason, and identify a specific target using only visual perception and language description. However, existing methods struggle with the memorization of complex spatial representations in aerial environments, reliable and interpretable action decision-making, and inefficient exploration and information gathering. To address these challenges, we introduce \textbf{APEX} (Aerial Parallel Explorer), a novel hierarchical agent designed for efficient exploration and target acquisition in complex aerial settings. APEX is built upon a modular, three-part architecture: 1) Dynamic Spatio-Semantic Mapping Memory, which leverages the zero-shot capability of a Vision-Language Model (VLM) to dynamically construct high-resolution 3D Attraction, Exploration, and Obstacle maps, serving as an interpretable memory mechanism. 2) Action Decision Module, trained with reinforcement learning, which translates this rich spatial understanding into a fine-grained and robust control policy. 3) Target Grounding Module, which employs an open-vocabulary detector to achieve definitive and generalizable target identification. All these components are integrated into a hierarchical, asynchronous, and parallel framework, effectively bypassing the VLM's inference latency and boosting the agent's proactivity in exploration. Extensive experiments show that APEX outperforms the previous state of the art by +4.2\% SR and +2.8\% SPL on challenging UAV-ON benchmarks, demonstrating its superior efficiency and the effectiveness of its hierarchical asynchronous design. Our source code is provided in \href{https://github.com/4amGodvzx/apex}{GitHub}.
\end{abstract}

\begin{figure}[ht]
    \centering
      \includegraphics[width=1.0\linewidth]{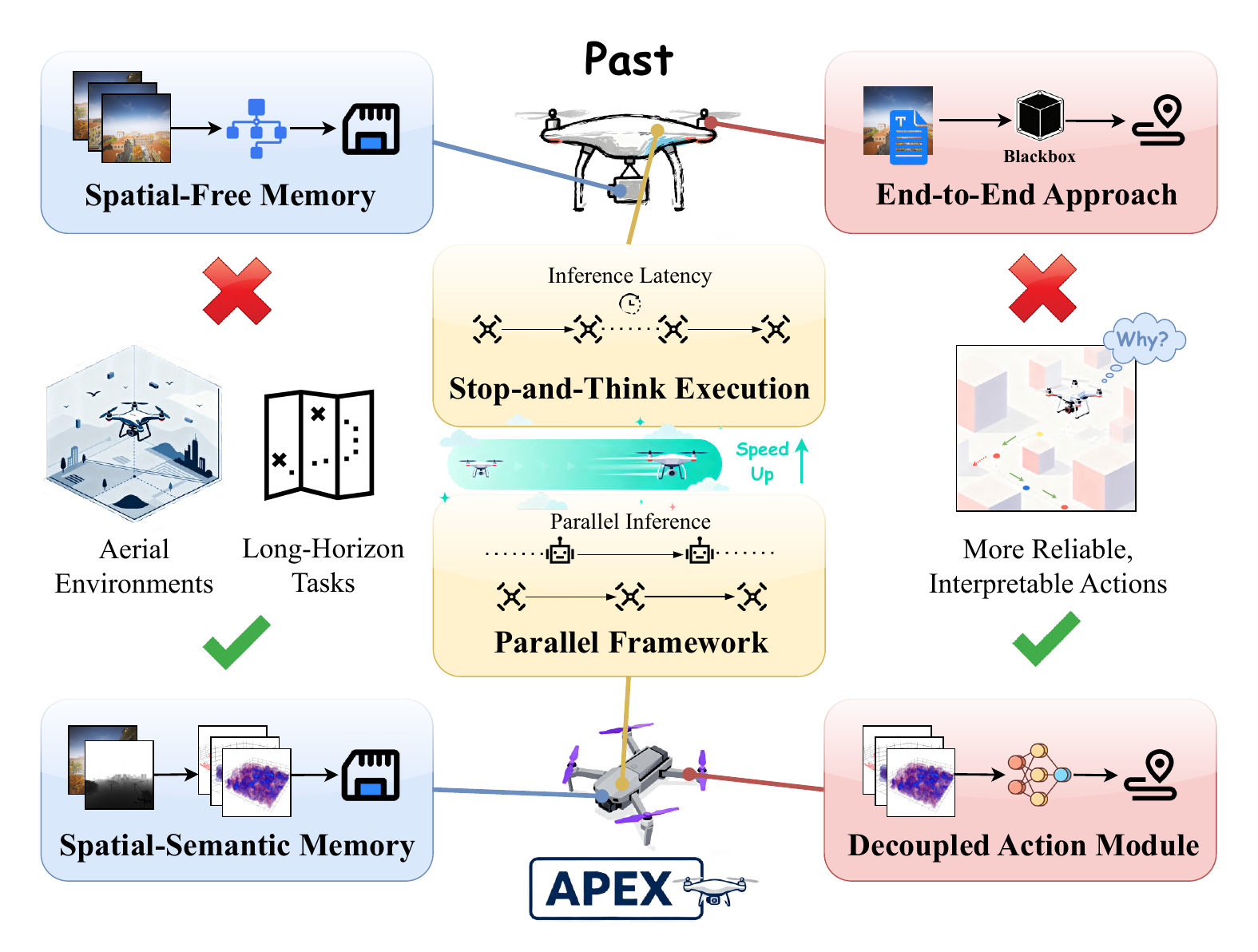}

      \caption{\textbf{APEX addresses three critical limitations in prior works}. 1) Memory: APEX employs a rich 3D dynamic map for robust spatial reasoning. 2) Decision-Making: APEX uses decoupled modules, separating semantic understanding from action generation for more reliable and interpretable control. 3) Efficiency: APEX  employs a parallel and asynchronous framework that mitigates inference latency and boosts exploration proactivity.}
      \label{fig:motivation}
    \end{figure}

\begin{figure*}[t]
  \centering
   \includegraphics[width=1.0\linewidth]{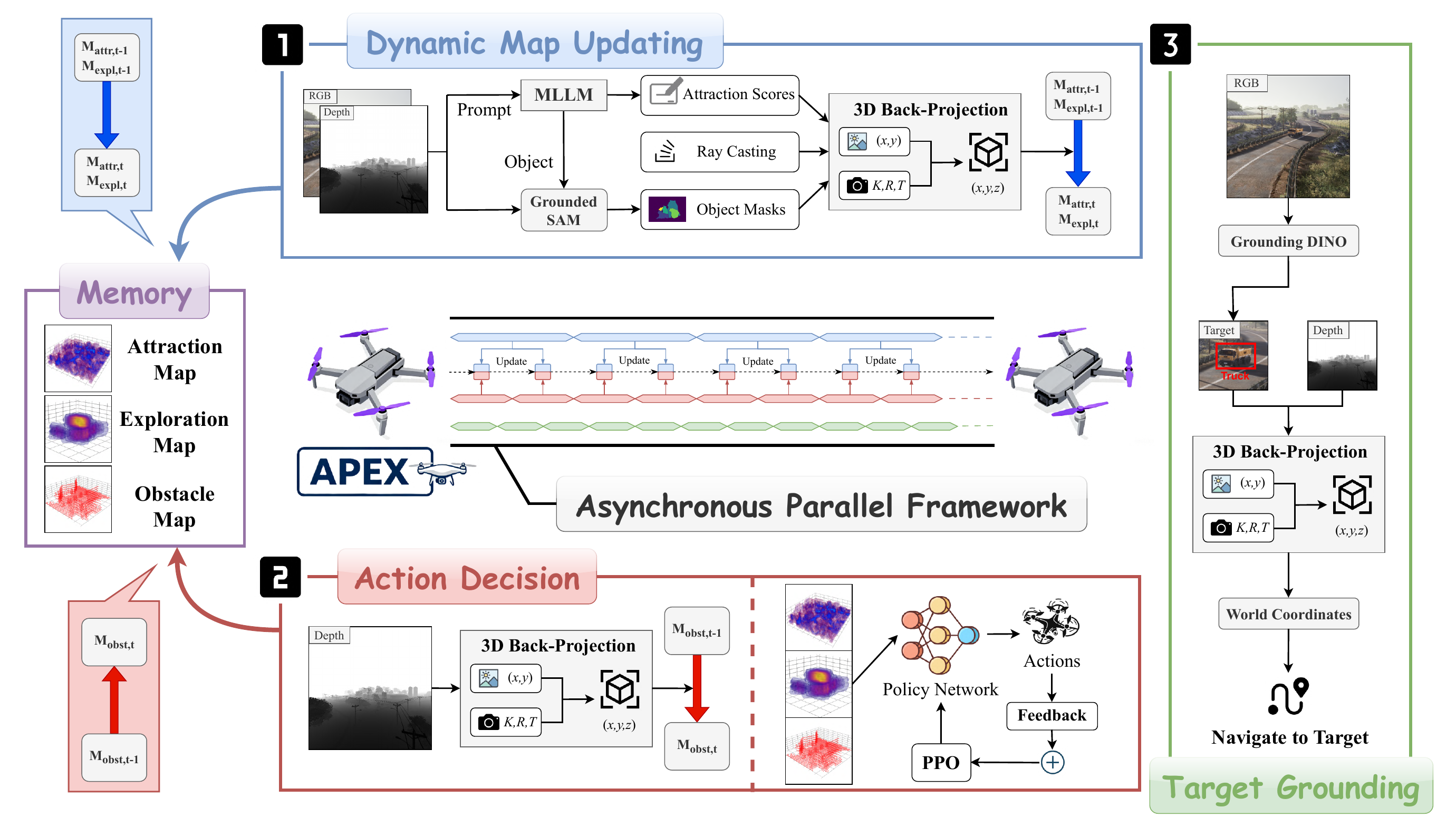}

   \caption{\textbf{The overall framework of APEX}.
   Module 1 extracts and integrates visual information into 3D dynamic spatio-semantic maps, serving as the agent's memory. Module 2 utilizes the stored information from Module 1 for action decision-making via an RL-based model. Module 3 continuously performs object detection on visual observations to obtain precise target locations. The three modules are integrated into a hierarchical asynchronous parallel system.}
   \label{fig:main}
\end{figure*}    
\section{Introduction}
\label{sec:intro}

The rapid increase in the use of Unmanned Aerial Vehicles (UAVs) \cite{Fanglong} and Vision-Language Models (VLMs) \cite{vlmsurvey,vlmsurvey2,vlmsurvey3} are driving advancements in aerial embodied intelligence \cite{uavllm,uavai}. The combination of these technologies has greatly advanced Vision-Language Navigation (VLN) \cite{vlnsurvey, vlnsurvey2,vlnsurvey3}, traditionally focused on indoor, ground-based scenarios, into the challenging domain of Aerial Vision-Language Navigation (AVLN) \cite{aerialvln,citynav, openfly,aeroverse}. This transition represents a significant leap in complexity, moving from constrained 2D planes to complex 3D spaces. However, a critical limitation of many existing AVLN frameworks is their reliance on fine-grained, step-by-step language instructions \cite{geotext,takingflightwithdialogue,aeroduo}. Such detailed guidance is often impractical to obtain in real-world settings and fundamentally constrains the agent's autonomy, limiting the scalability and applicability of these systems in open-ended exploration tasks \cite{logisticsvln,generalpurpose}.

To address this limitation, we focus on a more challenging and practical task: Aerial Object Goal Navigation (Aerial ObjectNav) \cite{uavon}. In this paradigm, the UAV agent is tasked with locating a specific object described by a high-level textual cue (\eg, ``a red tent'') using only its onboard visual sensors. This requires the agent to autonomously explore the environment, reason about spatial relationships, and actively search for the target, mirroring a more realistic operational scenario.

As shown in \cref{fig:motivation}, developing an Aerial ObjectNav agent faces three major challenges that are insufficiently addressed by prior work:

1) \textbf{Ineffective Spatial-Temporal Information Integration}. The transition from ground to aerial navigation brings an exponential increase in the volume and complexity of perceptual information \cite{airvista,selfprompting,refdrone}. Operating in expansive 3D environments, agents must execute long-horizon tasks that require effective long-term memory. However, many existing approaches \cite{flightgpt,avlngrid} fail to integrate historical observations with complex spatial information, resulting in inefficient exploration and repeated errors. Other works \cite{skyvln,navagent,geonav} employ abstract memory structures like topological maps, but these representations often sacrifice crucial geometric and metric fidelity, impairing the agent's ability to perform fine-grained reasoning and planning in complex aerial terrains.

2) \textbf{The Gap Between Semantic Understanding and Actionable Control}. While VLMs exhibit remarkable proficiency in grounding language to visual concepts, a significant gap exists between their high-level semantic reasoning and the low-level action control required for navigation \cite{vlasurvey}. Contemporary methods \cite{openuav,uavflow,racevla,cognitivedrone} attempt to use a VLM as an end-to-end decision-maker, directly mapping visual-linguistic inputs to actions. This approach not only demands vast quantities of high-quality demonstration data but also results in policies that are often unstable, unreliable, and lack interpretability, making them unsuitable for safety-critical applications, especially in aerial environments.

3) \textbf{Computational Latency and Real-Time Constraints}. The high computational cost of running large VLMs presents a critical challenge to real-world deployment \cite{ovib}. Most existing frameworks \cite{citynavagent,prpsearcher,uavon} overlook the severe time costs, implicitly assuming a ``stop-and-think'' execution model. This latency prevents the smooth, continuous motion necessary for efficient and practical UAV operation, making these systems unsuitable for real-time missions.

To overcome these challenges, we propose APEX (Aerial Parallel Explorer), a novel hierarchical agent designed for efficient aerial object navigation. APEX's core contribution lies in decoupling the complex navigation task into three specialized modules: a Dynamic Spatio-Semantic Mapping module, an RL-based Action Decision module, and a Target Grounding module for precise and definitive target identification. By integrating these components within an asynchronous, parallel framework, APEX enables effective long-horizon planning in complex aerial environments, enhances the reliability and interpretability of decision-making, and achieves high operational efficiency.

Before delving into details, our contributions are summarized as follows:
\begin{itemize}
    \item We propose a novel dynamic 3D mapping mechanism specifically for aerial object navigation, which enhances the agent's ability to integrate and maintain complex spatial information over long-horizon tasks in expansive 3D environments.
    \item We demonstrate the advantages of a modular, decoupled design for planning and control in Aerial ObjectNav. Through extensive experiments, we show that this approach surpasses end-to-end models in terms of reliability and interpretability.
    \item We introduce a multi-module, asynchronous parallel architecture. Our experiments demonstrate that this framework improves exploration efficiency in Aerial Object Navigation, which is often overlooked in prior work. This paradigm sets a new standard for building practical and responsive aerial agents.
\end{itemize}

\section{Related Work}
\label{sec:relatedwork}

\subsection{Object Goal Navigation}

Object Goal Navigation is an Embodied AI task where an agent must autonomously find a target object in an unseen environment \cite{embodiedagenteval}. Researches in this area, sped up by photorealistic simulators like Matterport3D \cite{matterport3d} and AI2-THOR \cite{ai2thor}, has evolved from classical modular pipelines to learning-based paradigms. To improve generalization, many modern methods employ imitation learning (IL) or reinforcement learning (RL) to train end-to-end navigation policies \cite{rlmethod,uavswarm,pirlnav,rapid}.

The capabilities of Multimodal Large Language Models (MLLMs) have recently revolutionized this domain by equipping agents with powerful commonsense reasoning \cite{mllmsurvey,mllmsurvey2}. For instance, NavGPT-2 \cite{navgpt2} aligns visual and linguistic information within a frozen LLM to bridge the performance gap between generalist LLM agents and specialized navigation models. Similarly, VoroNav \cite{voronav} leverages an LLM to interpret semantic and topological information from a dynamically map, enabling more efficient zero-shot exploration by reasoning about navigational waypoints.

However, the challenges of ObjectNav are significantly amplified when transitioning from ground to aerial navigation \cite{prpsearcher,raven}. The complexity of vast 3D environments places greater demands on an agent's perception and planning. To this end, UAV-ON \cite{uavon} established a large-scale simulation benchmark for aerial ObjectNav, while PRPSearcher \cite{prpsearcher} proposed a 3D semantic map-based framework to improve spatial understanding. Despite this progress, developing an agent that is both efficient and reliable for long-horizon aerial tasks remains a critical open problem. To address this, our Dynamic-Mapping Module improves the agent's abilities of memorizing in aerial navigation, while the Action Decision Module enables more reliable action outputs.

\subsection{Aerial Vision-Language Navigation}

Inspired by the advancements in ground-based VLN \cite{r2r,vlnce}, Aerial Vision-Language Navigation (AVLN) has recently emerged as a challenging new frontier, extending embodied navigation to the complexities of 3D aerial environments\cite{textguide,fela,vlfly}. Foundational works like AerialVLN \cite{aerialvln} provide a dedicated simulator, a large-scale dataset, and a baseline method. Following this, TravelUAV \cite{openuav} introduced continuous control for more realistic simulation and proposed an end-to-end LLM-based agent for aerial navigation. Recent improvements have focused on specific modules: To enhance long-horizon task performance, SkyVLN \cite{skyvln} and NavAgent \cite{navagent} introduced topological map architectures as the agent's memory modules, abstracting spatial relationships for efficient planning. To boost the generalization and adaptability of the VLM backbone, FlightGPT \cite{flightgpt} employs a two-stage training pipeline, including Supervised Finetuning and GRPO algorithm. To improve spatial reasoning, CityNavAgent \cite{citynavagent} enhanced the agent's understanding of 3D geometry by projecting visual information onto point clouds.

However, existing AVLN methods exhibit notable limitations. First, their performance is often contingent upon high-quality, detailed language instructions, which may not be available in more open-ended, goal-driven scenarios. Furthermore, a critical and under-explored issue is the loss of complex spatial information in their memory module designs. For instance, while topological maps are effective for high-level planning, they often discard the precise geometric details essential for fine-grained maneuvering and obstacle avoidance in cluttered 3D spaces. We address these problems with a system that navigates using simple goal descriptions, supported by dynamic 3D maps that retain crucial spatio-semantic information.


\section{Methodology}
\label{sec:method}

\begin{figure*}
  \centering
  \begin{subfigure}{0.52\linewidth}
    \includegraphics[width=1.0\linewidth]{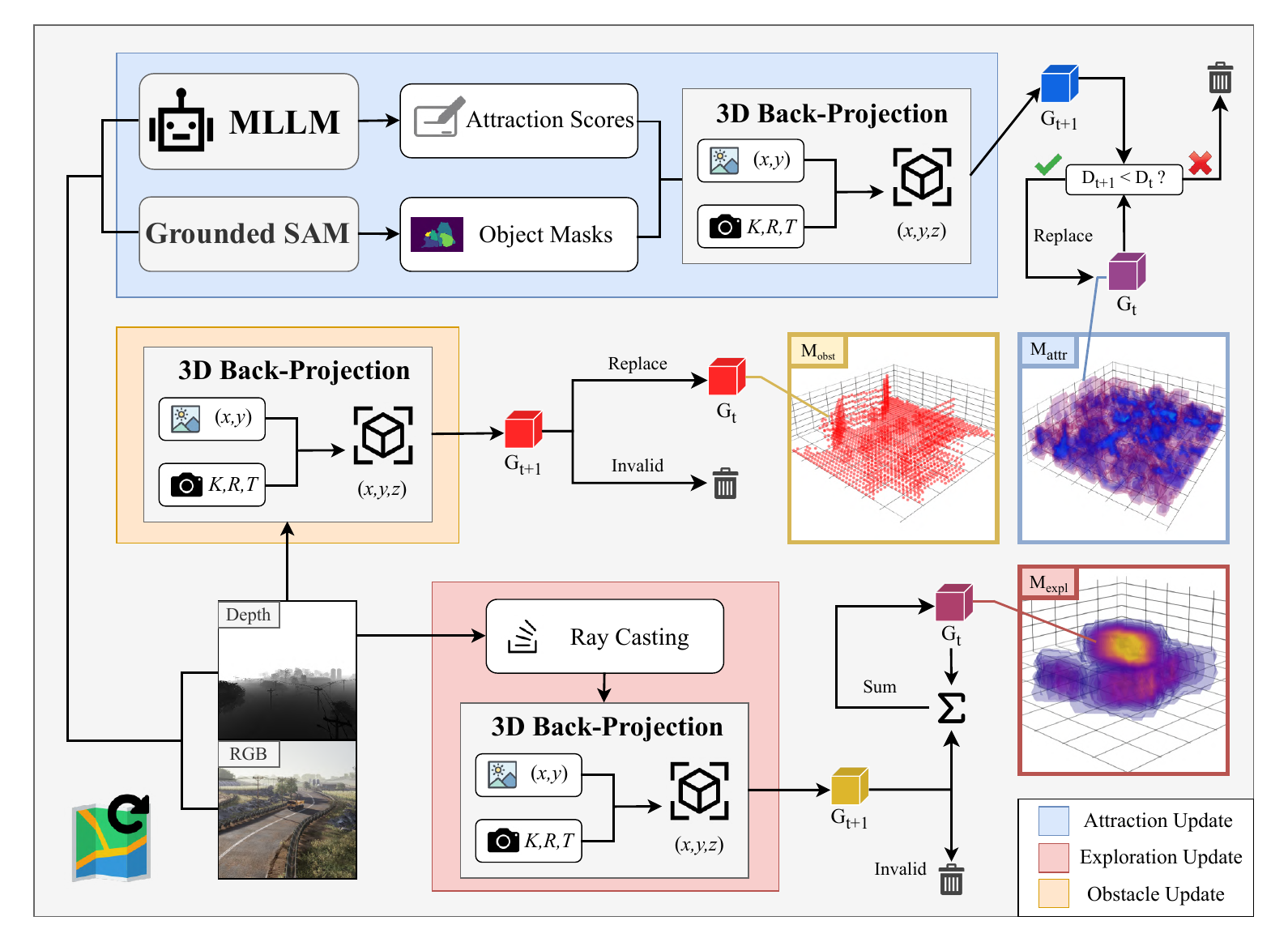}
    \caption{\textbf{3D Dynamic Map Updating}. The module processes RGB-D input to update three maps with specific rules: 1) Attraction Map: A grid's score is updated only if the current observation is geometrically closer. 2) Exploration Map: Scores from new observations are added to existing ones. 3) Obstacle Map: All grids with detected obstacles are marked as occupied.}
    \label{fig:short-a}
  \end{subfigure}
  \hfill
  \begin{subfigure}{0.46\linewidth}
    \includegraphics[width=1.0\linewidth]{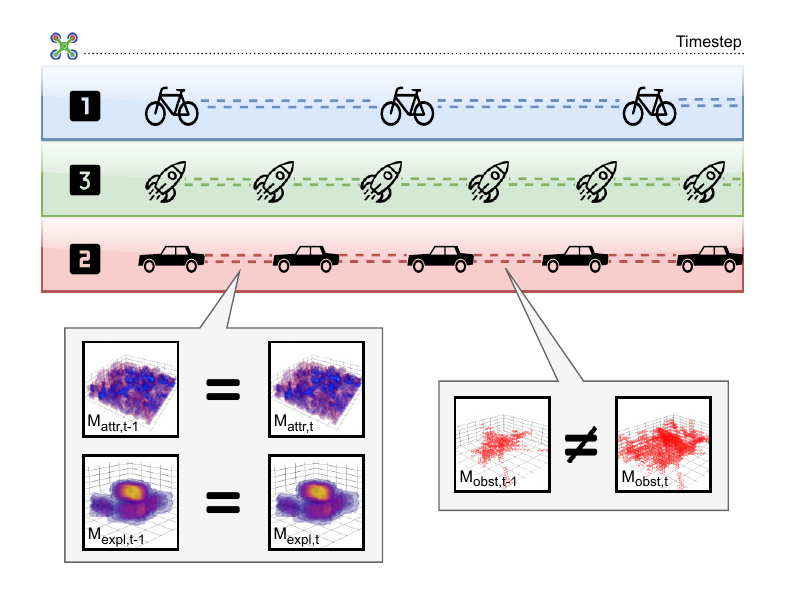}
    \caption{\textbf{Asynchronous Parallel Framework}. The agent's modules operate in parallel at different frequencies. Since the Map Update module runs slower than the Action Decision module, the Obstacle Map input for the policy is always updated between consecutive timesteps, while the Attraction and Exploration Maps may remain the same.}
    \label{fig:short-b}
  \end{subfigure}
  \caption{Detailed architecture of the Dynamic Map Module and the Asynchronous Parallel Framework.}
  \label{fig:short}
\end{figure*}

\subsection{Task Formulation}

We formulate the Aerial Object Goal Navigation task as a sequential decision-making process. At each timestep $t$, the agent receives a visual observation $O_t$, its self state $S_t$, and a text description $D$. In a basic setting, the agent's policy, denoted by $\pi$, maps the current sensory inputs and the goal description to an action $A_t$ from a discrete action space $\mathcal{A}$:
\begin{equation}
  A_t = \pi(O_t,S_t,D).
  \label{eq:pi}
\end{equation}
For long-horizon tasks that require efficient exploration and recall of the environment, we introduce a memory module $MEM$ to aggregate spatial-semantic information. The module updates the memory state from $MEM_{t-1}$ by incorporating new information from $O_t$ and $S_t$:
\begin{equation}
  MEM_t=f_{\mathrm{update}}(MEM_{t-1},O_t,S_t).
  \label{eq:mem}
\end{equation}
With this memory-augmented framework, the agent's policy now makes decisions on the accumulated memory:
\begin{equation}
  A_t = \pi(O_t,S_t,MEM_t,D).
  \label{eq:pi_m}
\end{equation}
The ultimate goal of the agent is to reach the target efficiently, formulated as finding a policy $\pi^*$ that maximizes the success metric.


\subsection{Overview}

As illustrated in \cref{fig:main}, the APEX architecture is designed to supports efficient Aerial ObjectNav by addressing three core challenges of the task. Each of them is handled by a dedicated module:

\noindent \textbf{Dynamic Spatio-Semantic Mapping Module}. To navigate purposefully, the agent first needs to build a model of the world. This module creates and maintains this model by processing visual and textual inputs from an MLLM \cite{qwen2.5vl} and a segmentation model \cite{groundedsam}. It generates three dynamic 3D maps: the Attraction Map to guide the agent toward the goal, the Obstacle Map for safety, and the Exploration Map to ensure active exploration.

\noindent \textbf{RL-based Action Decision Module}. With a clear understanding of the environment, the agent must then choose its next action. This module uses the generated maps as input for a Proximal Policy Optimization (PPO) \cite{ppo} policy network. The network learns to translate high-level spatial intelligence into low-level actions, enabling effective and reliable movements.

\noindent \textbf{Target Grounding Module}. To solve the ``last-mile'' problem of target identification, the agent needs to be certain that it has found the correct object. This module uses a dedicated object detector \cite{groundingdino} to confirm the target’s location. 

These modules operate in a parallel asynchronous framework, ensuring high efficiency and overcoming latency bottlenecks for smooth exploration. The algorithm flow is provided in Appendix A.

\subsection{3D Dynamic Grid-based Map}

As illustrated in \cref{fig:short-a}, our method utilizes a 3D grid-based map $M$ as a persistent memory carrier, which is dynamically updated with spatio-semantic information at each timestep. This process involves two key stages: 3D back-projection and map generation.

    \noindent \textbf{3D Back-projection}. Given a depth image, this process back-projects the 2D pixel grid into a 3D point cloud in the world frame. This transformation uses the camera intrinsics $K$, per-pixel depth values, and the current state of the agent $S_t$. Each resulting 3D point is then discretized to identify its corresponding grid index in the map $M$. We refer to this entire pipeline as the Reconstruction Projector, RP(·).

    \noindent \textbf{Attraction Map Generation}. For effective Object Goal Navigation, the agent must infer which regions are promising for finding the target object. To this end, we introduce the concept of an object-centric attraction map, $M_{\mathrm{attr}}$, a channel within our main map $M$. This map quantifies the semantic relevance of observed objects to the navigation goal. The generation process begins by leveraging MLLM's advanced capabilities in visual grounding and commonsense reasoning \cite{qwen2.5vl}. Given the current observation $O_t$ and the goal description $D$, the MLLM identifies $N$ distinct objects, outputting their text captions $c_1,c_2..., c_N$ and the corresponding attraction scores $c_1,c_2..., c_N$. A higher score $s_i$ indicates a stronger semantic link between the object $c_i$ and the target. We denote this process as CAP(·):
    \begin{equation}
      \{(c_i,s_i)\}_{i=1}^N= \mathrm{CAP}(O_t,D).
      \label{eq:cap}
    \end{equation}
    Subsequently, we employ an open-vocabulary segmentation model \cite{groundedsam} to obtain a pixel mask $Mask_i$ for each caption $c_i$, denoted as SEG(·):
    \begin{equation}
      {Mask_i}= \mathrm{SEG}(O_t,c_i).
      \label{eq:seg}
    \end{equation}
    Finally, the attraction map $M_{\mathrm{attr}}$ is updated by projecting these semantic masks into the 3D grid. For each voxel $v$ in the map, the update follows two rules:

    \begin{itemize}
      \item \textbf{Majority Ownership}: The voxel is claimed by the object $i^*$ that projects the most pixels into it.
      \item \textbf{Closest Observation First}: The voxel's score is updated only if the new observation is geometrically closer than any prior observation.
    \end{itemize}

    Let $s_\mathrm{new}$ and $d_{\mathrm{new}}$ be the attraction score and minimum depth of the pixels from the winning object $i$ that map to voxel $v$. The update rule for the attraction score $M_{\mathrm{attr}}(v)$ and its associated minimum depth $M_{\mathrm{depth}}(v)$ is:
    \begin{equation}
    \begin{cases}
      M_{\mathrm{attr}}(v) \gets s_{\mathrm{new}} \\
      M_{\mathrm{depth}}(v) \gets d_{\mathrm{new}}
    \end{cases}
    \text{ if } d_{\mathrm{new}} < M_{\mathrm{depth}}(v).
    \label{eq:attrup}
    \end{equation}
    This ensures that the map robustly represents the most relevant and reliable information from the agent's perspective.

    \noindent \textbf{Exploration Map Generation}. To incentivize the agent to explore unknown regions and gather more environmental information, we maintain an Exploration Map $M_{\mathrm{expl}}$. This channel quantifies the observation density of each spatial region, representing an ``exploration score''.

    The map is updated at each timestep $t$ using a ray casting approach. For the set of 3D world points $P_w$ from the current depth observation, we cast a set of rays $\mathcal{R}$ originating from the camera's position $t_t$ towards each point $P_w$. The set of all voxels traversed by a single ray $r$, denoted as $V(r)$, is identified using a 3D voxel traversal algorithm. The union of these sets constitutes all currently visible voxels:
    \begin{equation}
      \mathcal{V}_\mathrm{visible}=\bigcup_{r \in \{r\}} V(r).
      \label{eq:raycast}
    \end{equation}
    For each voxel $v \in \mathcal{V}_{\mathrm{visible}}$, we calculate an exploration gain that decays exponentially with its distance from the agent. The incremental update, $\Delta M_{\mathrm{expl}}$, is calculated as:
    \begin{equation}
      \Delta M_\mathrm{expl}(v)=\exp{(-\lambda \cdot \|p_v-t_t\|_2)},
      \label{eq:decay}
    \end{equation}
    where $p_v$ is the center of voxel $v$, and $\lambda$ is the decay rate. This incremental value is then accumulated into the map:
    \begin{equation}
      M_\mathrm{expl,t}(v) \gets M_\mathrm{expl,t-1}(v) + \Delta M_\mathrm{expl}(v).
      \label{eq:exploreup}
    \end{equation}
    This mechanism encourages the agent to prioritize navigating towards less observed regions.

    \noindent \textbf{Obstacle Map Generation}. To enhance safe navigation and path planning, we also construct an Obstacle Map, $M_{\mathrm{obst}}$. This map provides a persistent representation of occupied space. It is populated directly by the 3D Back-Projection process RP(·). Any voxel that contains one or more back-projected points from the depth map is marked as an obstacle. The update rule for a voxel $v$ is simply:
    \begin{equation}
      M_\mathrm{obst}(v) \gets 1 \quad \text{ if } {v} \text{ contains any } {P_w}.
      \label{eq:obstup}
    \end{equation}

\subsection{RL-based Action Decision Module}

    The Action Decision Module employs a policy network, trained with the Proximal Policy Optimization (PPO) algorithm, to map the constructed 3D maps to low-level control actions. This process includes feeding three distinct 3D maps into three separate CNN feature extractors. The resulting feature vectors are then concatenated into a unified representation, which is shared by both the actor and critic networks that form the core of the policy network. This map-based approach enhances the policy's reliability and interpretability compared to end-to-end methods. To inject navigation priors and improve training efficiency, we design a composite reward function, $R_{t}$, which is a weighted sum of several components:

    \noindent \textbf{Sparse Reward}. These rewards are given only at the end of an episode to signify ultimate success or failure.
    \begin{itemize}
        \item A large positive reward, $R_{\mathrm{success}}$, is given if the agent navigates to within a threshold distance of the target.
        \item A large negative reward, $R_{\mathrm{penalty}}$, is given for terminal states such as collisions or flying out of bounds.
    \end{itemize}
    
    \noindent \textbf{Attraction Reward}. This dense reward encourages the agent to leverage the semantic guidance from the Attraction Map. At each step $t$, the reward is calculated by summing the attraction scores of all observed voxels, weighted by a linear distance decay. For the set of visible voxels $\mathcal{V}_{\mathrm{visible}}$ within a distance threshold $d_{thresh}$, the reward is:
    \begin{equation}
      R_\mathrm{attr} = \sum_{v \in \mathcal{V}_{\mathrm{visible}}} M_{\mathrm{attr}}(v) \cdot \left(1 - \frac{\|p_v - t_t\|_2}{d_{\text{thresh}}}\right),
      \label{eq:rewardattr}
    \end{equation}
    where $M_{\mathrm{attr}}(v)$ is the attraction score of voxel $v$, $p_v$ is its world coordinate, $t_t$ is the agent's position at time $t$, and the term in parentheses provides the linear decay. This reward incentivizes the agent to move towards and observe regions with high semantic relevance to the target.
    
    \noindent \textbf{Exploration Reward}. This dense reward incentivizes the agent to explore unknown areas to gather more information. It is inversely proportional to the voxel's current exploration score in $M_{\mathrm{expl}}$:
    {
    \small
    \begin{equation}
      R_\mathrm{expl} = \sum_{v \in \mathcal{V}_{\mathrm{visible}}} (\epsilon-M_{\mathrm{expl}}(v)) \cdot \left(1 - \frac{\|p_v - t_t\|_2}{d_{\text{thresh}}}\right),
      \label{eq:rewardexpl}
    \end{equation}
    }where $\epsilon$ represents the saturation point for exploration.

    Finally, the total reward at each timestep $t$ is the weighted sum of these components:
    \begin{equation}
      R_t= R_\mathrm{attr} + \alpha R_\mathrm{expl} + R_\mathrm{spar}
      \label{eq:rewardtotal},
    \end{equation}where $\alpha$ balances the scale of the attraction reward and the exploration reward.

\subsection{Target Grounding Module}

    While the 3D map guides the agent to the general vicinity of the target, the Target Grounding Module is responsible for the final, precise localization. This module operates in parallel with other components, continuously scanning each visual observation to detect the target.

    At each step in this phase, the agent employs an open-vocabulary object detector, GD(·), on its current RGB observation $O_t$ to find instances matching the goal description $D$ \cite{groundingdino}. The detector outputs a set of candidate bounding boxes and their associated confidence scores:
    \begin{equation}
      \mathrm{(bbox_j,conf_j)}\}_{j=1}^K = \mathrm{GD}(O_t,D))
      \label{eq:groundingdino}.
    \end{equation}
    If the highest confidence score exceeds a predefined threshold, the target is considered successfully grounded. The precise 3D world coordinate of the target $P_{\mathrm{target}}$ will be computed by the Reconstruction Projector RP(·).


\subsection{Asynchronous Parallel Framework}

    To address the computational bottleneck imposed by large-scale VLM inference, we integrate our modules into an asynchronous parallel framework, which is illustrated in \cref{fig:short-b}. The Attraction Map and Exploration Map serve as shared data structures that mediate information transfer between different modules, which operate at varying frequencies. The framework consists of three main components, ordered from lowest to highest operational frequency:

    \noindent \textbf{Dynamic Map Module}. This module operates at a lower frequency and is responsible for the most computationally intensive tasks. It runs the VLM-based inference to generate attraction scores and performs ray casting to compute exploration gains. The resulting updates are asynchronously written to the shared Attraction Map and Exploration Map.

    \noindent \textbf{Action Decision Module}. Running at the agent's main control frequency, this module is responsible for real-time navigation. At each action step, it updates the Obstacle Map using the latest sensor data, without time-consuming processes. Then it reads the most recent versions of the Attraction Map and Exploration Map from the shared memory. These three maps collectively inform the action policy, ensureing that even if the semantic guidance is slightly stale, the agent can always navigate safely based on up-to-the-minute obstacle information.

    \noindent \textbf{Target Grounding Module}. This module also operates at a high frequency, continuously applying the open-vocabulary detector GD(·) to the RGB observations. This high-rate execution maximizes the probability of detecting the target as soon as it becomes visible, without being constrained by the slower semantic mapping cycle.

\section{Experiment}
\label{sec:experiment}

\begin{table*}
  \centering
  \caption{\textbf{Comparison with Baselines}. APEX achieves SOTA performance across various types of agents, including basic methods, ground navigation, and aerial navigation techniques.}
  \begin{tabular}{@{}c*{12}{c}@{}}
  \toprule
  \textbf{Method} & \multicolumn{4}{c}{\textbf{Seen}} & \multicolumn{4}{c}{\textbf{Unseen}} & \multicolumn{4}{c}{\textbf{Overall}}\\
  \cmidrule(lr){2-5} \cmidrule(lr){6-9} \cmidrule(lr){10-13}
  & \textbf{SR$\uparrow$} & \textbf{NE$\downarrow$} & \textbf{OSR$\uparrow$} & \textbf{SPL$\uparrow$} 
  & \textbf{SR$\uparrow$} & \textbf{NE$\downarrow$} & \textbf{OSR$\uparrow$} & \textbf{SPL$\uparrow$}
  & \textbf{SR$\uparrow$} & \textbf{NE$\downarrow$} & \textbf{OSR$\uparrow$} & \textbf{SPL$\uparrow$}  \\
  \midrule
  RE & 1.12 & 32.75 & 3.37 & 0.62 & 0.0 & \textbf{28.45} & 9.68 & 0.0 & 0.83 & 31.64 & 5.00 & 0.46 \\
  FBE~\cite{frontier} & 4.49 & 66.02 & 11.24 & 2.57 & 6.45 & 63.52 & 12.90 & 6.45 & 5.00 & 65.38 & 11.67 & 3.5 \\
  MLLM-N~\cite{qwen2.5vl} & 0.0 & \textbf{30.39} & 3.37 & 0.0 & 3.23 & 29.83 & 12.88 & 3.23 & 0.83 & \textbf{30.25} & 5.83 & 0.83 \\
  CLIP-H~\cite{uavon} & 6.74 & 46.38 & 12.36 & 4.58 & 3.23 & 49.51 & 16.13 & 3.67 & 5.83 & 47.19 & 13.33 & 4.34 \\
  L3MVN-Z~\cite{l3mvn} & 8.99 & 61.64 & 15.73 & 6.57 & 9.69 & 63.07 & 16.12 & 9.68 & 9.17 & 62.01 & 15.83 & 7.37 \\
  TRAVEL~\cite{openuav} & 6.74 & 56.66 & 13.48 & 4.82 & 9.68 & 53.12 & 16.15 & 7.60 & 7.48 & 55.75 & 14.17 & 5.54 \\
  AOA-F~\cite{uavon}  & 7.87 & 47.68 & 17.98 & 4.21 & 6.45 & 48.57 & 16.09 & 3.98 & 7.50 & 47.90 & 17.50 & 4.15 \\
  \textbf{APEX}  & \textbf{12.36} & 55.59 & \textbf{19.10} & \textbf{9.13} & \textbf{16.13} & 51.74 & \textbf{22.58} & \textbf{13.03} & \textbf{13.33} & 54.59 & \textbf{20.00} & \textbf{10.14} \\
  \bottomrule
  \end{tabular}
  \label{tab:tabm}
\end{table*}


\subsection{Experiment Settings}

    \noindent \textbf{Dataset and Environment}. We evaluate our proposed method using the UAV-ON benchmark \cite{uavon}. This recently proposed benchmark is specifically designed for the Aerial ObjectNav task, providing a challenging and realistic simulation environment. It features 14 photorealistic, large-scale outdoor environments and comprises a total of 10,000 navigation episodes, split for training and testing. In each episode, the agent is tasked with autonomously navigating to a target object based on a given textual description.

    \noindent \textbf{Evaluation Metrics}. We employ four standard metrics widely used in embodied navigation tasks to provide evaluations of navigation accuracy and path efficiency. Success Rate (SR) measures the percentage of episodes where the agent stops within a predefined distance threshold. Navigation Error (NE) is the average final Euclidean distance between the agent's position and the target's center at the end of each episode. Oracle Success Rate (OSR) measures success assuming an oracle stops the agent at the point along its trajectory that is closest to the target. Success-weighted Path (SPL) evaluates both success and efficiency by penalizing inefficient paths, weighting the success rate by the ratio of the optimal path length to the agent's actual path length.

    \noindent \textbf{Baselines}. To demonstrate the effectiveness of APEX, we compare it against a comprehensive set of baselines spanning different categories, including basic methods, ground navigation, and other aerial navigation techniques.
    \begin{itemize}
      \item \textbf{Random Exploration}: A naive agent that selects actions randomly at each timestep until the stop action is chosen.
      \item \textbf{Frontier-based Exploration} \cite{frontier}: A classic exploration method without leveraging semantic information.
      \item \textbf{MLLM-Direct Navigation} \cite{qwen2.5vl}: An end-to-end approach where MLLM directly maps raw egocentric observations and the target description to actions.
      \item \textbf{CLIP-H} \cite{uavon}: This method uses CLIP \cite{clip} to compute the similarity between observations and the textual goal, guiding the agent towards semantically relevant areas.
      \item \textbf{L3MVN-Zeroshot} \cite{l3mvn}: A novel method for ground-based ObjectNav. It uses an MLLM to fuse semantic information with a frontier map for navigation decisions.
      \item \textbf{TravelUAV} \cite{openuav}: A recent LLM-based framework designed for the AVLN task. To adapt to the experimental setup, the model's action outputs were discretized.
      \item \textbf{AOA-F} \cite{uavon}: An Aerial ObjectNav method that processes and feeds the target description, RGB observations, and depth observations into an LLM to generate actions.
    \end{itemize}

    \noindent \textbf{RL Training}. The policy network in our Action Decision Module maps 3D grid-based maps to discrete actions. We employ the Proximal Policy Optimization (PPO) algorithm for training this policy. PPO is widely used in navigation tasks because of its stability and strong performance \cite{ppo}, making it a reliable choice for our work. Our model is trained on a set of 36 tasks, comprising 4 distinct tasks across 9 environments. To accelerate training, we use 4 parallel environments, with each operating on a different map to enhance data diversity. Within each environment, the agent cycles through the map's designated tasks. For each task, the agent collects experience over 10 consecutive episodes before proceeding to the next.
    
    To enhance training stability, we adopt a two-stage training approach. Initially, a goal-agnostic exploration policy is pretrained using only the exploration reward $R_\mathrm{expl}$ and the terminal penalty $R_\mathrm{penalty}$. This helps the agent establish a robust foundation for safe navigation and obstacle avoidance. Subsequently, this model is fine-tuned by introducing the attraction reward $R_\mathrm{attr}$ and the success reward $R_\mathrm{success}$, guiding the policy to leverage semantic cues from the Attraction Map for efficient, goal-directed navigation. This curriculum effectively decouples the learning of general navigation skills from task-specific behaviors, fostering a final policy that balances exploration and exploitation. More training details are provided in Appendix D.
    
    \begin{table}[t]
      \centering
      \caption{Comparison of Efficiency and Security.}
      \begin{tabular}{ccc}
      \toprule
      \textbf{Method} & \textbf{Step Latency (s)} & \textbf{Safe Dist. (m)} \\
      \midrule
      L3MVN-Z & 1.26 & 330.64 \\
      TRAVEL & 1.71 & 223.17 \\
      AOA-F & 5.29 & 212.36 \\
      \textbf{APEX} & \textbf{0.97} & \textbf{345.51} \\
      \bottomrule
      \end{tabular}
      \label{tab:tabe}
    \end{table}
    
    \begin{figure*}[t]
    \centering
    \includegraphics[width=1.0\linewidth]{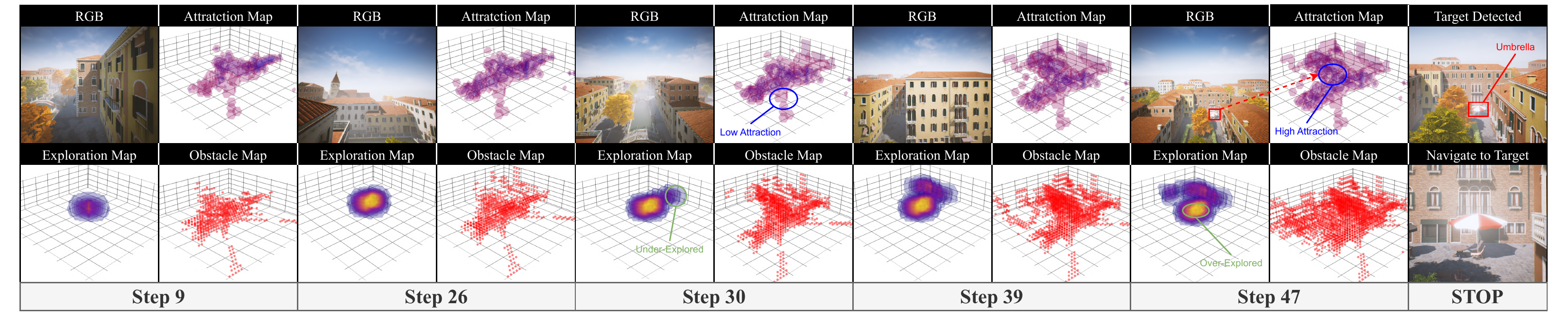}
    \caption{A case study of the APEX agent navigating to the target ``Umbrella''. The figure shows the dynamic integration of the Attraction, Exploration, and Obstacle Maps in the mission, leading to successful navigation. More visualized results are demonstrated in Appendix B.}
    \label{fig:case}
    \end{figure*}

\subsection{Experimental Results}

    \noindent \textbf{Comparison with Baselines}. \cref{tab:tabm} presents the comparative results of APEX and the baseline methods across four evaluation metrics. Among the basic methods, Frontier-based Exploration (FBE) significantly outperforms Random Exploration, demonstrating the fundamental benefit of systematic exploration over naive behavior. VLM-based methods, such as CLIP-H, highlight how the semantic understanding of visual scenes can effectively guide the agent. Similarly, TravelUAV, an AVLN method, also shows strong performance, but its framework is tailored for detailed linguistic instructions. Additionally, L3MVN, a method for ground-based ObjectNav, also delivers highly competitive results, underscoring the power of fusing semantic information with spatial maps. However, its performance is constrained by its design, which is not optimized for the complexities of 3D aerial environments. Our proposed method, APEX, establishes a new state-of-the-art on this benchmark, achieving a 4.16\% improvement in Success Rate (SR) and a 2.77\% improvement in Success-weighted Path Length (SPL), while also attaining the highest Oracle Success Rate (OSR). Regarding the higher Navigation Error (NE), we attribute this to the agent's proactive exploration strategy, which can lead to a greater final distance from the target in the subset of unsuccessful episodes.
    
    \noindent \textbf{Efficiency and Security}. \cref{tab:tabe} compares the operational efficiency and safety of APEX against three top-performing baselines. Our method achieves the lowest step latency (the time between consecutive actions), demonstrating that our Asynchronous Parallel Framework effectively boosts exploration efficiency. Furthermore, APEX attains the highest average safe distance, a measure of collision-free exploration capability, showing the enhanced navigation reliability provided by our RL-based Action Decision Module.

\subsection{Ablation Study}
    \begin{table}[t] 
      \centering
      \caption{Effect of the attraction-exploration trade-off parameter $\alpha$.}
      \begin{tabular}{ccccc}
      \toprule
      $\alpha$ & SR$\uparrow$ & NE$\downarrow$ & OSR$\uparrow$ & SPL$\uparrow$ \\
      \midrule
      0.05 & 5.83 & \textbf{29.11} & 9.17 & 5.28  \\
      0.1 & 9.17 & 41.92 & 15.83 & 8.23  \\
      \textbf{0.2} & \textbf{13.33} & 54.59 & \textbf{20.00} & 10.14 \\
      0.5 & 11.67 & 58.55 & 15.83 & 9.65  \\
      0.8 & 12.50 & 62.27 & 16.67 & \textbf{11.45}  \\
      1.0 & 7.50 & 66.44 & 13.33 & 4.85  \\
      \bottomrule
      \end{tabular}
      \label{tab:tabh}
    \end{table}
    To validate the contributions of key components in APEX, we conduct a series of ablation studies.
    
    First, we investigate the impact of the reward balancing parameter $\alpha$ in our RL-based Action Decision Module. This parameter balances the agent's tendency between exploitation and exploration (\cref{eq:rewardtotal}). As shown in \cref{tab:tabh}, we evaluate various values for $\alpha$. The results indicate that an $\alpha$ of 0.2 achieves the best overall performance, striking an optimal balance between attraction and exploration rewards.

    Next, we analyze the necessity of each core module, with results summarized in \cref{tab:taba}. To verify the function of the 3D Dynamic Grid-based Map, we removed it and used raw RGB and depth images as direct observational inputs to the Action Decision Module. The model struggled to converge under this setting and achieved poor navigation performance. To assess the impact of the RL-based Action Decision Module, we replaced it with a reward-prediction-based heuristic algorithm. The details of this algorithm can be seen in Appendix C. Although this heuristic achieved acceptable results, its tendency to get trapped in local optima led to performance that was inferior to our original module. Finally, to demonstrate the importance of the Target Grounding Module, we replaced it with a simple rule to navigate towards the highest activation point in the attraction map, which led to a substantial drop in performance and underscored the necessity of our specialized mechanism for accurate target localization.
        \begin{table}[t] 
      \caption{Ablation Study of Core Modules.}
      \centering
      \begin{tabular}{ccccc}
      \toprule
      Method & SR$\uparrow$ & NE$\downarrow$ & OSR$\uparrow$ & SPL$\uparrow$ \\
      \midrule
      w/o 3D-Map & 1.67 & 45.62 & 4.17 & 0.73 \\
      w/o RL-AD & 12.50 & \textbf{42.08} & 19.17 & 10.11  \\
      w/o TG & 5.00 & 48.33 & 9.17 & 4.17  \\
      \textbf{APEX} & \textbf{13.33} & 54.59 & \textbf{20.00} & \textbf{10.14}  \\
      \bottomrule
      \end{tabular}
      \label{tab:taba}
    \end{table}
\section{Conclusion}
\label{sec:conclusion}

In this study, we introduced APEX (Aerial Parallel Explorer), a novel agent for Aerial Object Goal Navigation. APEX features a hierarchical architecture consisting of three specialized modules: Dynamic Spatio-Semantic Mapping Module, RL-based Decision Making module, and Target Grounding module. These components are efficiently integrated through a novel Asynchronous Parallel Framework. Experimental results on the UAV-ON Benchmark demonstrate that APEX not only achieves state-of-the-art performance but also exhibits superior reliability and higher operational efficiency. In the future, we will conduct further research and optimization, including fine-tuning pre-trained VLM and MLLM models specifically for the Aerial ObjectNav task, as well as comparing the effects of different novel RL methods on APEX.

{
    \small
    \bibliographystyle{ieeenat_fullname}
    \bibliography{main}
}

\clearpage
\setcounter{page}{1}
\maketitlesupplementary

\section{Algorithm Flow}
\label{sec:algorithmflow}
\cref{alg:flow} and \cref{alg:mapupdate} illustrates the overall workflow of our agent. We employ a multi-process architecture to enable parallel execution of the three modules.
\begin{algorithm}[htbp]
    \caption{Overall Workflow}
    \label{alg:flow}

    \textbf{Input:} Observations $O$ \\
    \textbf{Input:} States $S$ \\
    \textbf{Input:} Target Description $D$ \\

    \begin{itemize}
        \item $MEM$ denotes the shared information, including $M_\text{attr}$, $M_\text{expl}$, $M_\text{obst}$, and others.
    \end{itemize}

    \begin{algorithmic}[1]
        \Statex
        \Function{Map Updating}{$MEM$,$O$,$S$,$D$}
            \While{not IsDetected}
                \State $c,s \gets \text{CAP}(O,D)$
                \State $\mathcal{M} \gets \text{SEG}(O,c)$
                \State $MEM \gets \text{Update}(MEM,\mathcal{M},s,O,S)$ 
            \EndWhile
            \State process.join()
        \EndFunction
        \Statex
        \Function{Action Decision}{$MEM$,$S$,$O$,$p$}
            \While{True}
                \If{IsDetected}
                    \State $\text{Navigate}(p)$
                \Else
                    \State \textbf{break}
                \EndIf
                \State $MEM \gets \text{Update}(MEM,O,S)$
                \State $a \gets \text{Policy}(MEM,S)$
                \State $\text{Execute}(a)$
            \EndWhile
            \State process.join()
        \EndFunction
        \Statex
        \Function{Target Grounding}{}
            \While{not IsDetected}
                \State $b \gets \text{GD}(O,D)$
                \State $p \gets \text{RP}(b,O,S)$
            \EndWhile
            \State process.join()
        \EndFunction
    \end{algorithmic}
\end{algorithm}

\begin{algorithm}[htbp]
    \caption{Memory Updating Workflow}
    \label{alg:mapupdate}
    \textbf{Input:} Masks $\mathcal{M}$ \\
    \textbf{Input:} Observations $O$ \\
    \textbf{Input:} States $S$ \\
    \textbf{Input:} Attraction Scores $s$ \\
    \begin{algorithmic}[1]
        \Function{AttrUpdate}{$M_{\text{attr}}, s, \mathcal{M}, O, S$}
            \State $\mathcal{G}_{\text{vis}}$: the set of visible grid cells in the current view.
            \For{$g \in \mathcal{G}_{\text{vis}}$}
                \State $P_g$: the set of pixels from $O$ that project to $g$.
                \State $Mask^* \gets \arg\max_{Mask_i \in \mathcal{M}} |P_g \cap Mask_i|$
                \If{$Mask^*$ is found}
                    \State $s^* \gets \text{score in } s \text{ corresponding to } Mask^*$
                    \State $d_{\text{new}} \gets \text{Dis}(S.\text{pos}, g.\text{center})$
                    \If{$d_{\text{new}} < M_{\text{attr}}[g].\text{dis}$}
                        \State $M_{\text{attr}}[g].\text{score} \gets s^*$
                        \State $M_{\text{attr}}[g].\text{dis} \gets d_{\text{new}}$
                    \EndIf
                \EndIf
            \EndFor
            \State \Return $M_{\text{attr}}$
        \EndFunction
        \Statex
        \Function{ExplUpdate}{$M_{\text{expl}}, O, S$}
            \State $\mathcal{P}_{\text{end}}$: the set of 3D world points from $O$.
            \State $\mathcal{G}_{\text{traversed}} \gets \text{TraceRay}(S.\text{pos}, \mathcal{P}_{\text{end}})$
            \For{$g \in \mathcal{G}_{\text{traversed}}$}
                \State $d \gets \text{distance}(S.\text{pos}, g.\text{center})$
                \State $v_{\text{current}} \gets M_{\text{expl}}[g]$
                \State $v_{\text{new}} \gets \text{ExplRate}(d, v_{\text{current}})$
                \State $M_{\text{expl}}[g] \gets v_{\text{current}} + v_{\text{new}}$
            \EndFor
            \State \Return $M_{\text{expl}}$
        \EndFunction
        \Statex
        \Function{ObstUpdate}{$M_{\text{obst}}, O, S$}
            \State $\mathcal{G}_{\text{depth}}$: the set of grid cells hit by depth rays from $O$.
            \State $M_{\text{obst}}[g] \gets 1, \quad \forall g \in \mathcal{G}_{\text{depth}}$
            \State \Return $M_{\text{obst}}$
        \EndFunction
        \Statex
        \Function{TraceRay}{$S_{\text{start}}, \mathcal{P}_{\text{end}}$}
            \State $\mathcal{G}_{\text{traversed}}$: the set of grid cells
            \For{$p_{\text{end}} \in \mathcal{P}_{\text{end}}$}
                \State $\vec{r} = p_{\text{end}} - S_{\text{start}}$
                \State Sample points $\mathcal{P}_s$ along $\vec{r}$ at regular intervals.
                \State $\mathcal{G}_{\text{traversed}} \gets \text{RP}(\mathcal{P}_s)$
            \EndFor
            \State \Return $\mathcal{G}_{\text{traversed}}$
        \EndFunction
    \end{algorithmic}
\end{algorithm}

\section{More Visualization Results}
\label{sec:morevisual}

\subsection{Cases}

\begin{figure*}[ht]
    \centering
    \includegraphics[width=1.0\linewidth]{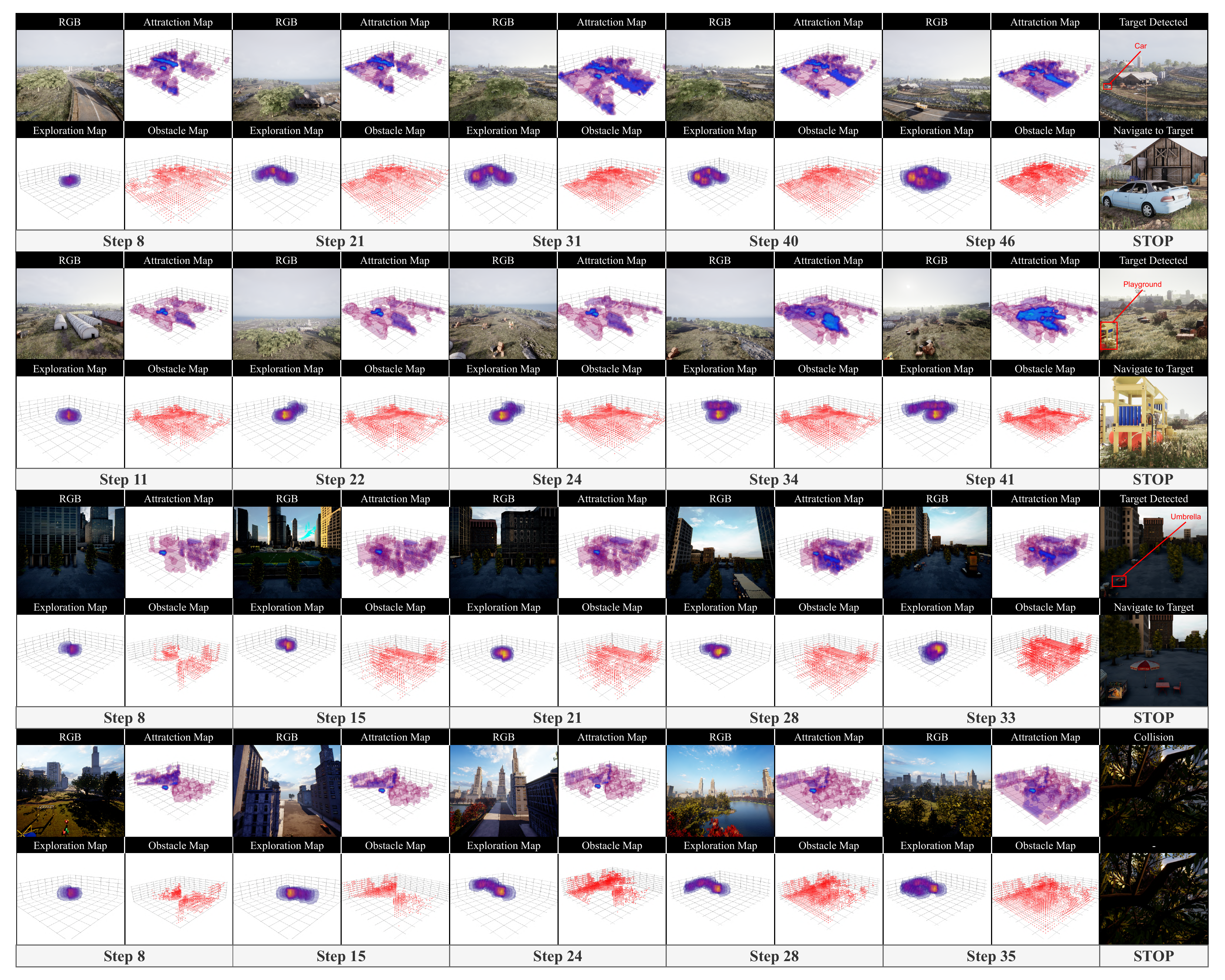}
    \caption{More visualized results. The first three examples are successful cases, with the targets "Car", "Playground", and "Umbrella" respectively. The last one is a failure case, with the target being "Sandbox".}
    \label{fig:mcase}
\end{figure*}

We provide additional visualizations to further illustrate APEX's performance. \cref{fig:mcase} shows four navigation episodes: three successful searches for "Car", "Playground", and "Umbrella", and one failure case targeting a "Sandbox". The failure was caused by a collision with a tree, which suggests a potential limitation in our Obstacle Map's ability to detect sparse obstacles. 
\subsection{Trajectories}
\begin{figure*}[ht]
    \centering
    \includegraphics[width=1.0\linewidth]{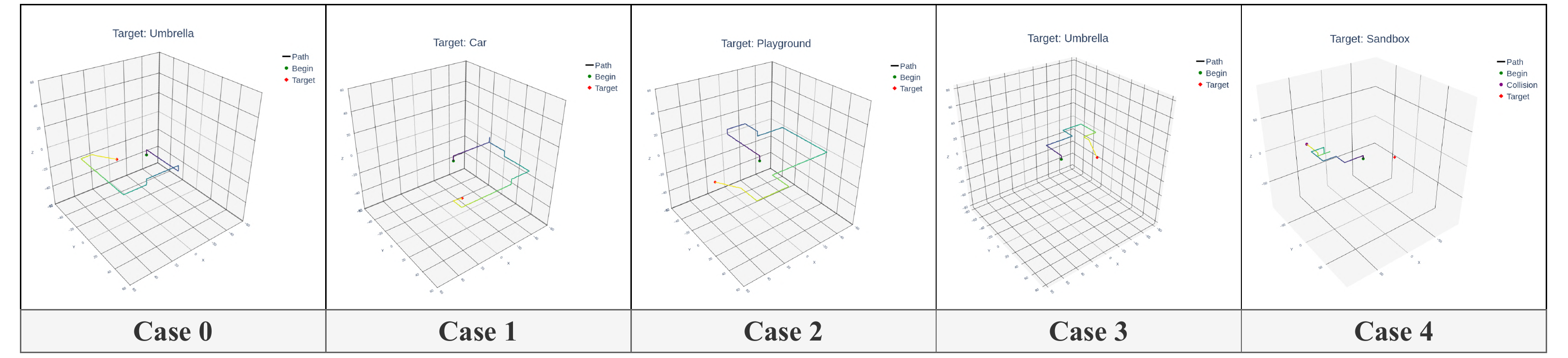}
    \caption{Complete trajectories of visualization examples.}
    \label{fig:path}
\end{figure*}

\cref{fig:path} demonstrates the complete trajectories for these four cases (labeled as Case 1-4), in addition to the main paper's example (Case 0).

\section{Details of Heuristic Algorithm}
\label{heuristic}
The heuristic algorithm mentioned in Ablation Study is detailed in \cref{alg:heuristic}. The core idea is to simulate each possible action, evaluate the potential reward based on attraction and exploration criteria, and select the action with the highest predicted reward.

\begin{algorithm}[htbp]
    \caption{Heuristic Action Decision}
    \label{alg:heuristic}

    \textbf{Input:} Attraction Map $M_\text{attr}$ \\
    \textbf{Input:} Exploration Map $M_\text{expl}$ \\
    \textbf{Input:} Obstacle Map $M_\text{obst}$ \\
    \textbf{Input:} Current State $s$ \\
    \begin{itemize}
        \item Let $a$ be an action from the action set $\mathcal{A}$.
        \item Let $s'$ be the hypothetical state after action $a$.
        \item The \textbf{attraction reward} $R_{\text{attr}}$ and \textbf{exploration reward} $R_{\text{expl}}$ is defined in main paper.
    \end{itemize}
    \bigskip
    \begin{algorithmic}[1]
        \State Initialize $R$ to store the predicted reward.
        \For{$a \in \mathcal{A}$}
            \State $s' \gets \text{Simulate}(a, s, M_\text{attr},M_\text{expl},M_\text{obst})$
            \If{$\text{IsUnsafe}(s')$}
                \State $R[a] \gets -\infty$
                \State \textbf{continue}
            \Else
                \State $R_[a]\gets W_{\text{attr}} R_{\text{attr}}(s') + W_{\text{expl}} R_{\text{expl}}(s')$
            \EndIf
        \EndFor
        \State \Return $\arg\max_{a \in A} R[a]$
    \end{algorithmic}
\end{algorithm}

\section{More Training Details}
\label{moretraining}

\subsection{Offline Map Generation}
\begin{figure*}[!t]
    \centering
    \includegraphics[width=1.0\linewidth]{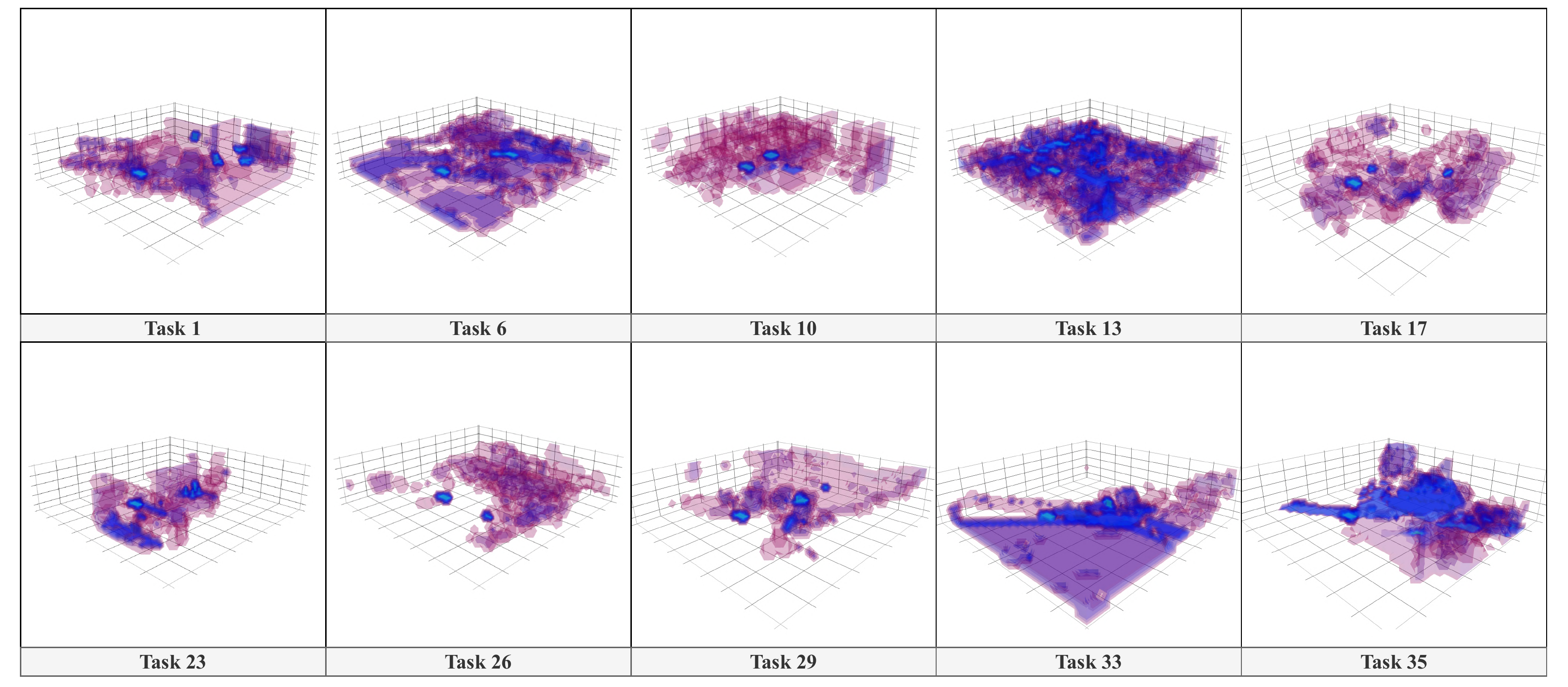}
    \caption{Examples of collected attraction maps.}
    \label{fig:data}
\end{figure*}

Directly incorporating the VLM for attraction map generation within the reinforcement learning loop is infeasible due to its high inference latency. To address this, we pre-generate a complete attraction map for each of the 36 tasks before training begins, which are shown in \cref{fig:data}.

For each task, we designed a flight path to cover the entire relevant search area. An agent was simulated along this path to generate and store a full attraction map. During RL training, instead of calling the VLM, the agent identifies its currently visible grid cells via the depth map and populates its local attraction map by looking up the pre-computed values. This method bypasses the VLM's latency, making the RL training process computationally available.
\begin{figure}[h]
    \centering
    \includegraphics[width=1.0\linewidth]{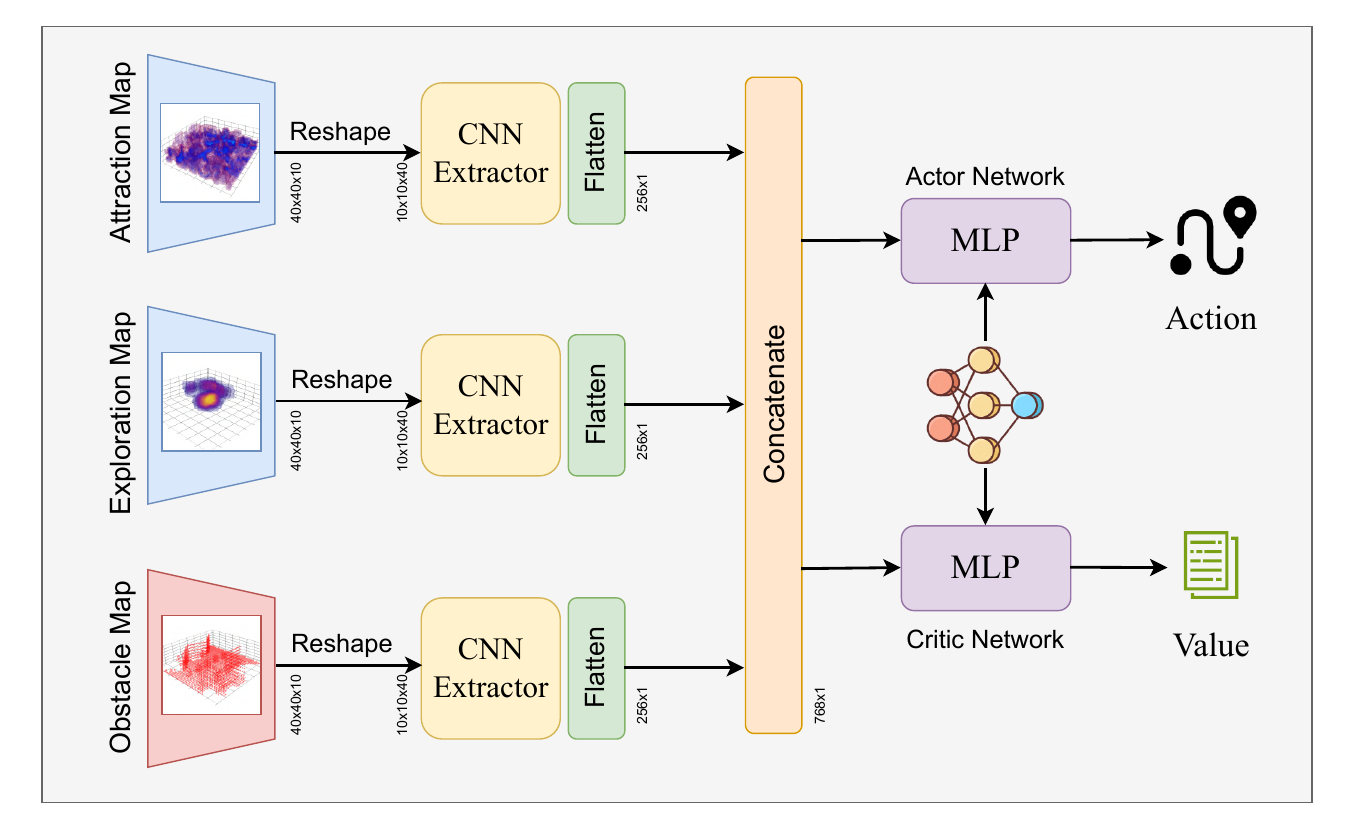}
    \caption{The architecture of policy network.}
    \label{fig:policy}
\end{figure}

\subsection{Policy Network Architecture}

\cref{fig:policy} illustrates the architecture of the policy network in our RL-based Action Decision Module. It utilizes three CNNs to extract features from the Attraction, Exploration, and Obstacle maps, respectively. To process the 3D map data, the z-axis is treated as the channel dimension for the convolutional layers. The resulting feature vectors are concatenated and shared between an Actor (MLP) and a Critic (MLP) head. The Actor network outputs the final action policy. This implementation is based on the MultiInputPolicy architecture provided by the Stable-Baselines3 \cite{sb3}.

\end{document}